\documentclass{article}

\usepackage[square,comma,sort&compress,numbers]{natbib}

\usepackage[preprint]{neurips_2021}

\usepackage[utf8]{inputenc} 
\usepackage[T1]{fontenc}    
\usepackage{hyperref}       
\usepackage{url}            
\usepackage{booktabs}       
\usepackage{amsfonts}       
\usepackage{nicefrac}       
\usepackage{microtype}      
\usepackage{xcolor}         
\usepackage{bm}
\usepackage{comment}
\usepackage{pgfplots}
\usepackage{makecell}
\usepackage{multicol,multirow}
\usepackage{amsmath,amssymb}

\pgfplotsset{width=7cm,compat=1.13}

\title{Modeling Text-visual Mutual Dependency for Multi-modal dialog Generation}

\author{
Shuhe Wang$^\spadesuit$, 
Yuxian Meng$^\clubsuit$, 
Xiaofei Sun$^\clubsuit$, Fei Wu$^\blacklozenge$\\
{\bf Rongbin Ouyang$^\spadesuit$, Rui Yan$^\bigstar$,
Tianwei Zhang$^\blacktriangle$, Jiwei Li$^{\blacklozenge\clubsuit}$}\\
$^\clubsuit$ Shannon.AI, $^\spadesuit$Peking University \\
$^\blacklozenge$Zhejiang University, $^\bigstar$Renmin University of China, $^\blacktriangle$Nanyang Technological University \\ 
  \{yuxian\_meng, xiaofei\_sun, jiwei\_li\}@shannonai.com\\
  wangshuhe@stu.pku.edu.cn, ouyang@pku.edu.cn\\
  wufei@zju.edu.cn, ruiyan@ruc.edu.cn, tianwei.zhang@ntu.edu.sg
}

\begin{document}

\maketitle

\begin{abstract}
Multi-modal dialog modeling is of growing interest. In this work, we propose frameworks to resolve a specific case of multi-modal dialog generation that better mimics multi-modal dialog generation in the real world, where each dialog turn is associated with the  visual context in which it takes place. Specifically, we propose to model the mutual dependency between text-visual features, where the model not only needs to learn the  probability of generating the next dialog utterance given preceding dialog utterances and visual contexts,  but also the  probability of predicting the visual features in which a dialog utterance takes place, leading the generated dialog utterance specific to the visual context. We observe significant performance boosts over vanilla models when the mutual dependency between text and visual features is modeled.\footnote{Code is available at \url{https://github.com/ShannonAI/OpenViDial}.}
\end{abstract}

\section{Introduction}
Multi-modal learning 
is of growing interest in  recent years \cite{das2017visual,gan2020large,radford2021learning,jia2021scaling}, and jointly modeling multiple modalities has shown notable effectiveness in improving models' ability 
to understand 
 visual and textual semantics, such as image captioning \cite{xia2020xgpt,cho2021unifying}, visual question answering \cite{chen2019uniter,li2020oscar,gan2020large} and text-to-image generation \cite{ramesh2021zero,radford2021learning}. 
As an important subfield of multi-modal learning, multi-modal dialog generation, targets generating coherent and informative dialog utterances specific to the visual contexts.
Although a few existing works have proposed to employ state-of-the-art multi-modal  models for multi-modal dialog generation \cite{shuster2018image,cui2019user,shuster2020multi}, they mainly focus on question-answering style dialog generation  grounded in a single image, rather than each image per dialog turn. This learning paradigm limits the application scope of  multi-modal dialog generation models on real-world scenarios where
dialog takes place in  visual contexts that change over time.


In this work , we propose frameworks to resolve a specific case of multi-modal dialog generation that 
better mimics multi-modal  dialog generation in the real world,  
where each dialog turn is associated with the  visual context in which  it takes place. 
Specifically, we first propose  
vanilla visual  
 models to extract and incorporate the visual features into sequence-to-sequence dialog generation, where each model extracts visual features at a different level: from using only textual features to using
 coarse-grained
  image-level features, and to the fine-grained object-level features.
Further, we propose to model the mutual dependency between textual and visual features,  
where 
the dialog model not only needs to learn the  probability of generating the next utterance given preceding dialog utterances and visual contexts, but also to model the backward probability of predicting visual features given the dialog utterance, leading 
the generated 
 dialog utterance specific to the visual context.

We conduct extensive experiments on the OpenViDial dataset \cite{meng2020openvidial}, and  experimental results show that incorporating visual features at a fine-grained granularity outperforms models that do not use visual features or only use coarse-grained visual features. 
Further, we observe significant performance boosts over vanilla 
visual 
 models when 
the  mutual visual-text dependencies are modeled, exhibiting 
its necessity  when  time-varying visual contexts needed to be considered.
The proposed models can act as strong baselines for future related works.

\section{Related Work}
\subsection{Textual Dialog Generation}
Existing works on building reliable dialog systems are generally divided into two categories: chit-chat open-domain dialog generation \cite{weizenbaum1966eliza,Wallace2009,li2017teaching} and task-oriented dialog generation \cite{young2013pomdp}.
Attempts to open-domain dialog generation include generating more coherent \cite{li2016deep,li2017adversarial,adiwardana2020towards}, diverse \cite{xu2018dp,baheti2018generating}, personalized \cite{li2016persona,mesgar-etal-2021-improving} utterances. 
With the emergence of task-oriented datasets \cite{shah2018building,wen2016network,budzianowski2018multiwoz,eric2019multiwoz}, more practice has been devoted to task-oriented dialog generation, which usually involves a pipeline of intent classification \cite{shi2016deep}, dialog state tracking \cite{henderson2013deep,henderson2014second,henderson2014word}, dialog policy making \cite{cuayahuitl2015strategic,li2017end} and dialog generation \cite{dhingra2016towards}. Dialog state tracking, due to its importance in bridging the user's intent and certain dialog states, has gained numerous attention over recent years \cite{mrkvsic2016neural,eric2017copy,lei2018sequicity,gao-etal-2019-dialog,wu2019alternating,hosseini2020simple,andreas2020task,li2021zero}. The prevalence of pretraining on large-scale unlabeled corpora also spurs a wealth of dialog generation systems under both the open-domain and task-oriented settings \cite{mehri2019pretraining,zhang2019dialogpt,roller2020recipes,huang2020challenges,gu2020tailored,lin2021m6}, leading to new SOTA results on dialog benchmarks.

\subsection{Jointly Modeling Visual and Textual Information}
Multimodal models have proven their ability of modeling interactions between different modalities and better undertanding the semantics behind textual utterances \cite{das2017learning,seo2017visual,lu2017best,mostafazadeh2017image,yang2019making,schwartz2019factor,gan2019multi,niu2019recursive,kang2019dual}, and pretraining on additional data gives further performance boosts for a variety of established vision-and-language tasks, such as visual question answering \cite{lu2019vilbert,alberti2019fusion,li2020oscar}, visual commonsense reasoning \cite{li2019unicodervl,li2019visualbert} and text-to-image generation \cite{lin2021m6,ramesh2021zero,radford2021learning}.
However, these works focus on the QA style visual dialog, rather than the conversation style with which we are more concerned. 
Another strand of works address multi-turn dialog generation grounded with vision \cite{saha2018towards,liao2018knowledge,alamri2018audio,cui2019user,firdaus2020multidm}. 
\cite{mostafazadeh2017image} studied the task of image-grounded conversations where utterances are generated about a shared image.
\cite{shuster2018image} constructed Image-Chat, a collection of consecutive turns by two interlocutors about an image along with their style traits. They used residual networks \cite{kaiming2016resnet} to encode images and Transformers to encode texts, and observed performance enhancement when incorporating visual information.  
\cite{shuster2020multi} investigated combining different state-of-the-art dialog agents and vision models for multimodal dialog generation. By carefully selecting components and training strategies, the best model surpasses existing multimodal systems regarding both automated and engagingness metrics.
Different from aforementioned works, we base this work on the intuition that the underlying semantics behind an image and a piece of text referring to the same event should be highly correlated, and we are motivated to study the effectiveness of mutual information between vision and text features in multi-modal dialog generation.

\section{The OpenVidial Dataset and Task Statement}
Different from previous datasets that focus on limited-domain multi-modal conversations such as 
E-commerce 
\cite{saha2018towards,firdaus2020multidm} or communications grounded in a single image \cite{shuster2018image}, the recently released large open-domain multi-modal dataset OpenViDial \cite{meng2020openvidial} contains millions of 
dialog turns, with each dialog turn associated with its specific visual context (image), in which the dialog utterance takes place, rather than a single image for the whole episode.
This better mimics multi-modal dialog generation in the real world.  
Regarding the dialog episode, the average number of turns for each episode in OpenViDial is 14, which is the largest among existing open-domain multi-modal dialog datasets. Therefore, in this work, we use the OpenViDial dataset 
as the benchmark for designing and testing different multi-modal dialog generation models.  

To be more detailed, OpenViDial consists of a set of dialog episodes
 $(X,Z)\in\mathcal{D}$, where $X=\{x_1,\cdots,x_n\}$ is a sequence of dialog turns 
in texts
and $Z=\{z_1,\cdots,z_n\}$ is a sequence of images.
Each dialog turn $x_n$ and the corresponding image $z_n$ are paired, 
and
 $z_n$ is the visual context which $x_n$ takes place in. 
$n$ denotes the length of the dialog episode. 
Each dialog turn $x_j (1\le j\le n)$ is a sequence of tokens, where
 $x_j = \{w_{j,1}, w_{j,2}, ..., w_{j,n_j} \}$, and $n_j$ denotes the length of the text turn $x_j$. 
We use $h_{j,k}$ to represent the word representation of token $w_{j,k}$. 

The task of multi-modal dialog generation in the OpenViDial style is to generate the next dialog turn $x_{j+1}$ conditioning on the preceding dialog turns $x_{\le j}$, the preceding visual contexts $z_{\le j}$ and the current visual context $z_{j+1}$, in an autoregressive manner:
\begin{equation}
   p(x_{j+1}|x_{<j},z_{\le j+1})= \sum_{k=1}^{k=n_{j+1}}p(w_{j+1,k}|x_{<j},z_{\le j+1},w_{j+1,<k})
\end{equation}
We introduce three families of models to tackle this problem.

\begin{figure*}
    \centering
    \includegraphics[scale=0.5]{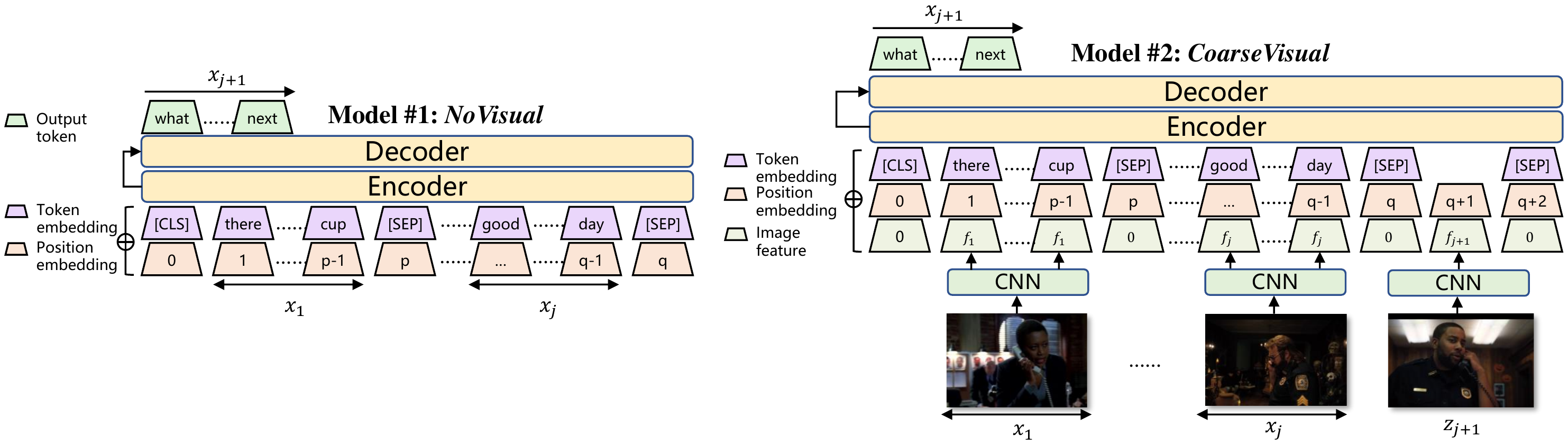}
    \includegraphics[scale=0.53]{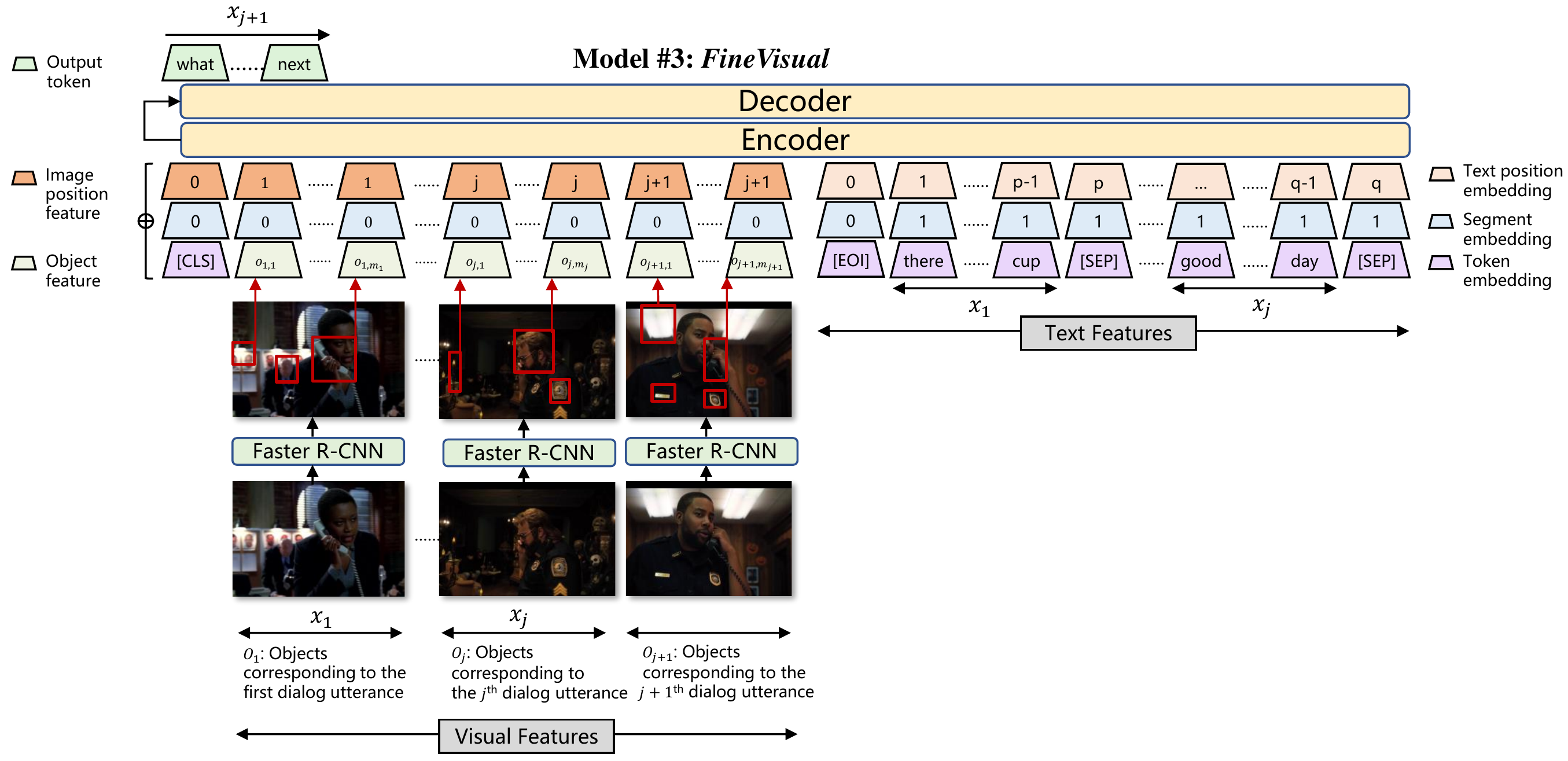}
    \caption{An overview of the proposed models NoVisual, CoarseVisual and FineVisual.}
    \label{fig:overview}
\end{figure*}

\section{Vanilla Visual Dialog Models} 
\label{section:visualDialogModels}

In this section, we introduce three visual dialog models, as shown in Figure \ref{fig:overview}. These models use both textual and visual contexts and employ the self-attention mechanism \cite{vaswani2017transformer} to model their interplay. The granularity of the visual features ranges from coarse-grained image features extracted from CNNs \cite{he2016deep} to fine-grained object features extracted from Faster R-CNNs \cite{ren2015faster}, each of which represents visual information at different levels.

\subsection{The {NoVisual} (NV) Model}
We first introduce a model that uses only dialog texts without visual information, where the model degenerates to a uni-modal dialog generation model. The model is optimized to minimize the following negative log-likelihood (NLL) loss:
\begin{equation}
  \mathcal{L}_{\text{NoVisual}}=-\sum_{(X,Z)\in\mathcal{D}}\sum_{j=0}^{n-1}p(x_{j+1}| x_{\le j} )
\end{equation}
We use a standard Transformer architecture \cite{vaswani2017transformer} as the model backbone. 
The numbers of encoder layer and decoder layer are both 3, with 8 heads in each layer and input dimension of 512.
We pack the all preceding dialog histories $x_{\le j}$ into a long sequence with a spacial \texttt{[SEP]} token as the delimiter between two
consecutive
 dialog turns. 
Sentence-positional and
token-positional 
positional embeddings are  added to word representations, which are fed to the transformer as inputs. 

\subsection{The {CoarseVisual} (CV) Model}
Our second model employs a naive approach to inject visual information into dialog generation, which we refer to as CoarseVisual (CV). More concretely, we first use a ResNet-50 model \cite{he2016deep}
pre-trained on  ImageNet \cite{5206848}
 to extract a high-dimensional feature $f_j$ for image $z_j$.
Then, 
for all tokens $w_{j,k}$ in the $j$-th dialog utterance, 
we add the image feature $f_j$ to its word representation $h_{j,k}$, forming the input layer representation $h^0_{j,k}$ as the input to the dialog model:
\begin{equation}
h^0_{j,k} = h_{j,k} + f_j
\end{equation}
The concatenation of all input token representations for the $j$-th dialog utterance is denoted by: 
\begin{equation}
h^0_j=[h^0_{j,1},\cdots,h^0_{j,n_j}]
\end{equation}
Hence, the input to the encoder
 is  given by $\{\texttt{[CLS]},h^0_1,\texttt{[SEP]},h^0_2,\texttt{[SEP]},\cdots,h^0_j,\texttt{[SEP]},f_{j+1},\texttt{[SEP]}\}$. 
$f_{j+1}$ represents the encoded feature of image $z_{j+1}$. The {CoarseVisual} model is then trained  to predict 
the forthcoming dialog utterance $x_{j+1}$ by minimizing the  NLL loss:
\begin{equation}
  \mathcal{L}_{\text{CoarseVisual}}=-\sum_{(X,Z)\in\mathcal{D}}\sum_{j=0}^{n-1}p(x_{j+1}|x_{\le j} , f_{\le j+1})
\end{equation}

\subsection{The {FineVisual} (FV) Model}
While the {CoarseVisual} model is able to combine the vision and text modalities, it performs at a coarse level for extracting 
global image features. This might be insufficient to model fine-grained visual elements in images such as facial expressions, body gestures as well as physical motions. Hence, we use Faster R-CNN \cite{ren2015faster}  pretrained on Visual Genome \cite{krishna2017visual} to extract fine-grained visual semantic objects. 
For an input image $z_j$, Faster R-CNN returns a set of detected objects in the image, each of which is captured by a dense feature representation. Let
 $O_j=\{o_{j,1},\cdots, o_{j,q}, \cdots, o_{j,m_j}\}$ denote the set of object features for image $z_j$, where $m_j$ is the number of extracted objects. 
 Each extracted feature
 can be mapped back to a bounding box / region (i.e., Region-of-Interest (RoI))
 in the original image. 
 For each dialog turn $x_{j+1}$ to generate, the input to the model is  $\{\texttt{[CLS]},O_1,\cdots,O_{j+1},\texttt{[EOI]},x_1,\texttt{[SEP]},\cdots,x_j,\texttt{[SEP]}\}$. \texttt{[EOI]} is a special end-of-image token denoting the end of the sequence of object features. 
Similar to the {CoarseVisual} model, the {FineVisual} model is optimized to minimize the following NLL loss:
\begin{equation}
  \mathcal{L}_{\text{FineVisual}}=-\sum_{(X,Z)\in\mathcal{D}}\sum_{j=0}^{n-1}p(x_{j+1}|x_{\le j},O_{\le j+1})
\end{equation}
For visual features $\{O_1,\cdots,O_{j+1},\texttt{[EOI]}\}$, an image-specific
positional feature highlighting objects across different images is added to object representations.

\section{Modeling Visual-Text Mutual Dependency } \label{section:MI}
We find that sometimes, the CV and FV models still suffering from generating text utterances that are not very related or even unrelated to the visual contexts. 
This is due to the nature of objective of  generative modeling, i.e., $p(x_{j+1}|x_{\le j} , f_{\le j+1})$: 
though the generation $p(x_{j+1})$ is conditioned on the preceding visual features $f_{\le j+1}$, there is no guarantee on whether or how much 
the evidence in $f_{\le j+1}$ is used. 
To strengthen the connection between visual features and text features, and enforce the model to generate utterances that are {\bf very} specific to its visual contexts,
we propose to incorporate the backward probability of generating features of visual contexts 
 given text utterances.
 The model is trained to learn the mutual information (MI) between visual contexts and text features, 
  as will be described below. 

\subsection{MI-CV}
For the CV model, the text utterance to generate  has the largest 
combination of the forward
probability, i.e., generating the current text utterance given preceding dialog utterances and visual contexts $p(x_{j+1}|x_{\le j},f_{\le j+1})$, and 
the backward probability, 
i.e.,  predicting the visual features in which a dialog utterance takes place $p(f_{j+1}|x_{j+1})$: 
\begin{equation}
\begin{split}
  \hat{x}_{j+1}=\mathop{\arg\max}_{x_{j+1}}\{(1-\lambda)\log p(x_{j+1}|x_{\le j},f_{\le j+1})
  +\lambda\log p(f_{j+1}|x_{j+1})\}
  \label{eq:MI-cv1}
\end{split}
\end{equation}

where $\lambda\in(0,1)$ is a hyperparameter that controls the trade-off between the forward and  the backward probabilities. 
We use negative sampling  
to empirically compute $\log p(f_{j+1}|x_{j+1})$.
Specifically,  
we  use a light discriminative network to approximate this probability. We first concatenate the high dimensional visual feature $f_{j+1}$ to the textual feature $t_{j+1,k}$ of each token in $x_{j+1}$. The textual feature $t_{j+1,k}$ is produced by the model encoder. Then we feed the resulting sequence of high dimensional features into a single-layer feed forward network (FFN) of dimensionality of 512 followed by the sigmoid function to output the likelihood $q(f_{j+1},t_{j+1,k})$ for each token. Last, we obtain $\log q(f_{j+1},x_{j+1})$ by averaging all these token-level log-likelihoods:
\begin{equation}
  \log q(f_{j+1},x_{j+1})=\frac{1}{n_{j+1}}\sum_{k=1}^{n_{j+1}}\log q(f_{j+1},t_{j+1,k})
\end{equation}
We treat $\log q(f_{j+1},x_{j+1})$ as an approximate of $\log p(f_{j+1}|x_{j+1})$.
To train the FNN, we randomly draw an image feature $f_{l}(l\not=j+1)$ to form negative examples for each positive $(x_{j+1},f_{j+1})$ example and minimize the following loss:
\begin{equation}
\begin{split}
  \mathcal{L}=-\sum_{(X,Z)\in\mathcal{D}}\sum_{j=0}^{n-1}\{\log q(f_{j+1},x_{j+1})
  -\sum_l \log q(f_{l,l\neq j+1},x_{j+1})\}
\end{split}
\label{hah}
\end{equation}

Another issue with Eq.(\ref{eq:MI-cv1}) is that it is infeasible to iterative over all possible $x_{j+1}$ to search for the optimal one as the search space grows exponentially large with respect to the length of the utterance $x_{j+1}$. To reduce the search space, we build an N-best list $\mathcal{N}$ which consists of the $N$ dialog utterances with the highest probabilities decoded using the forward probability $\log p(x_{j+1}|x_{\le j},f_{\le j+1})$, and then 
rerank the N-best list
 with the highest MI-interpolated probability in Eq.(\ref{eq:MI-cv1}). As it is a common practice to model the mutual dependency between the  dialog utterance $x_{j+1}$ and the prior dialog utterance $x_{j}$, we also incorporate the probability of generating $x_{j}$ given  $x_{j+1}$, i.e., $p(x_{j}|x_{j+1})$ into Eq.(\ref{eq:MI-cv1}) to build semantic connections between consecutive dialogs.
Combining all, we can rewrite Eq.(\ref{eq:MI-cv1}) by the following equation:
\begin{equation}
\begin{split}
  \hat{x}_{j+1}=\mathop{\arg\max}_{x_{j+1}\in\mathcal{N}}\{\lambda_{1}\log p(x_{j+1}|x_{\le j},f_{\le j+1})
  +\lambda_{2}\log p(x_{j}|x_{j+1})
  +\lambda_{3}\log q(f_{j+1},x_{j+1}) \}
\end{split}
\end{equation}
where $\lambda_{i=1,2,3}$ are three hyperparameters satisfying $\sum_{i=1}^{3}\lambda_{i}=1$.
We use two Transformer models respectively for $p(x_{j+1}|x_{\le j},f_{\le j+1})$ and $ p(x_{j}|x_{j+1})$. Both models have 3 encoder layers and 3 decode layers, 
with 8 heads in each layer and an input dimensionality of 512.

\subsection{MI-FV}
Different from the CV model, the FineVisual model is learning to generate the forthcoming dialog utterance $x_{j+1}$ based on fine-grained object features. 
To adjust the MI structure to the FV model, 
we propose to predict the fine-grained object features given the forthcoming dialog utterance: 
\begin{equation}
\begin{split}
  \hat{x}_{j+1}=\mathop{\arg\max}_{x_{j+1}\in\mathcal{N}}\{&\lambda_{1}\log p(x_{j+1}|x_{\le j}, O_{\le j+1})
  +\lambda_{2}\log p(x_{j}|x_{j+1})
  +\lambda_{3}\log q_2(O_{j+1},x_{j+1}) \}
\end{split}
\end{equation}
where $\lambda_{i=1,2,3}$ are hyperparameters satisfying $\sum_{i=1}^{3}\lambda_{i}=1$. 
For the term $\log q_2(O_{j+1},x_{j+1})$, we first use dimension-wise mean pooling to compress the set of extracted features $O_{j+1}$ into one high dimensional feature $o_{j+1}$. Then we  concatenate it to every high dimensional representation $t_{j+1, k}$ produced by the encoder, forming a sequence of features as input to an FFN of dimensionality 512. Similar to what we do in MI-CV, we can get $\log q_2(O_{j+1},x_{j+1})$ by averaging all the log-likelihoods $\log q_2(o_{j+1},t_{j+1,k})$ produced by the FFN model with the sigmoid function:
\begin{equation}
  \log q_2(O_{j+1},x_{j+1})=\frac{1}{n_{j+1}}\sum_{k=1}^{n_{j+1}}\log q_2(o_{j+1},t_{j+1,k})
\end{equation}
Again, we train the model by randomly sampling negative examples and  minimizing the loss:
\begin{equation}
\begin{split}
  \mathcal{L}=-\sum_{(X,Z)\in\mathcal{D}}\sum_{j=0}^{n-1}\{\log q_2(O_{j+1},x_{j+1})
  -\sum_l \log q_2(O_{l,l\neq j+1},x_{j+1})\}
\end{split}
\end{equation}

\section{Experiments}
We train all models using the Adam \cite{kingma2014adam} optimizer and decay the learning rate (LR) based on the inverse square root of the update number after the step of Warmup, which we set to 6000. For MI, the size of the N-best list $\mathcal{N}$ is set to 5.
We also apply MI to the NV model as a baseline. The model is denoted by MI+NV.

\begin{table}[!t]
  \small
  \centering
  \scalebox{0.75}{
  \begin{tabular}{ccccccccccc}\toprule
  \multicolumn{1}{c}{\bf Model} & 
  \multicolumn{1}{c}{\bf BLEU-1} & 
  \multicolumn{1}{c}{\bf BLEU-2} & 
  \multicolumn{1}{c}{\bf BLEU-4} & 
  \multicolumn{1}{c}{\bf Dis-1} & 
  \multicolumn{1}{c}{\bf Dis-2} & 
  \multicolumn{1}{c}{\bf Dis-3} & 
  \multicolumn{1}{c}{\bf Dis-4} &
  \multicolumn{1}{c}{\bf ROUGE-1} & 
  \multicolumn{1}{c}{\bf ROUGE-2} & 
  \multicolumn{1}{c}{\bf ROUGE-4}
  \\\midrule 
  {\it NV w/o MI} & \text{14.06} & \text{3.80} & \text{0.95} & \text{0.0006} & \text{0.0019} & \text{0.0031} & \text{0.0043} & \text{0.06787} & \text{0.01464} & \text{0.00224}   \\\hline
  {\it CV w/o MI} & \text{14.70} & \text{4.38} & \text{1.14} & \text{0.0023} & \text{0.0090} & \text{0.0178} & \text{0.0272} & \text{0.08773} & \text{0.02067} & \textbf{0.00347}  \\
  & (+4.6\%) & (+15.3\%) & (+20.0\%) & (+283\%) & (+374\%) &(+474\%) & (+533\%) & (+29.3\%) &(+41.2\%) & (+54.9\%) \\\hline
  {\it FV w/o MI} & \textbf{14.85} & \textbf{4.61} & \textbf{1.19} & \textbf{0.0026} & \textbf{0.0112} & \textbf{0.0246} & \textbf{0.0406} & \textbf{0.09083} & \textbf{0.02085} & \text{0.00329}  \\
  & (+5.6\%) & (+21.3\%) & (+25.3\%) & (+333\%) & (+489\%) &(+694\%) & (+844\%) & (+33.8\%) &(+42.4\%) & (+46.9\%) \\
  \bottomrule
  \end{tabular}
  }
  \caption{Automatic evaluation results for vanilla models on the OpenViDial dataset. }
  \label{tab:automatic_0}
\end{table}

\begin{table}[!t]
  \small
  \centering
  \scalebox{0.7}{
  \begin{tabular}{l|ccccccccccc}\toprule
  \multicolumn{1}{l}{\bf System} & 
  \multicolumn{1}{c}{\bf Model} & 
  \multicolumn{1}{c}{\bf BLEU-1} & 
  \multicolumn{1}{c}{\bf BLEU-2} & 
  \multicolumn{1}{c}{\bf BLEU-4} & 
  \multicolumn{1}{c}{\bf Dis-1} & 
  \multicolumn{1}{c}{\bf Dis-2} & 
  \multicolumn{1}{c}{\bf Dis-3} & 
  \multicolumn{1}{c}{\bf Dis-4} &
  \multicolumn{1}{c}{\bf ROUGE-1} & 
  \multicolumn{1}{c}{\bf ROUGE-2} & 
  \multicolumn{1}{c}{\bf ROUGE-4}
  \\\midrule 
  \multirow{3}{*}{\bf NV} & {\it w/o MI} & \text{14.06} & \text{3.80} & \text{0.95} & \text{0.0006} & \text{0.0019} & \text{0.0031} & \text{0.0043} & \text{0.06787} & \text{0.01464} & \text{0.00224} \\
  & \multirow{2}{*}{\it w/ MI}& \textbf{14.27} & \textbf{3.89} & \textbf{0.99} & \textbf{0.0006} & \textbf{0.0022} & \textbf{0.0036} & \textbf{0.0050} & \textbf{0.06918} & \textbf{0.01497} & \textbf{0.00238} \\
  & & (+1.49\%) & (+2.37\%) & (+4.21\%) & (+0.00\%) & (+15.79\%) &(+16.13\%) & (+16.28\%) & (+1.93\%) &(+2.25\%) & (+6.25\%) \\\hline
  \multirow{3}{*}{\bf CV} & {\it w/o MI} & \text{14.70} & \text{4.38} & \text{1.14} & \text{0.0023} & \text{0.0090} & \text{0.0178} & \text{0.0272} & \text{0.08773} & \text{0.02067} & \text{0.00347} \\
  & \multirow{2}{*}{\it w/ MI} & \textbf{14.77} & \textbf{4.46} & \textbf{1.16} & \textbf{0.0023} & \textbf{0.0091} & \textbf{0.0181} & \textbf{0.0276} & \textbf{0.08791} & \textbf{0.02077} & \textbf{0.00350} \\
  & & (+0.48\%) & (+1.83\%) & (+1.75\%) & (+0.00\%) & (+1.11\%) &(+1.69\%) & (+1.47\%) & (+0.21\%) &(+0.48\%) & (+0.86\%) \\\hline
  \multirow{3}{*}{\bf FV} & {\it w/o MI} & \text{14.85} & \text{4.61} & \text{1.19} & \text{0.0026} & \text{0.0112} & \text{0.0246} & \text{0.0406} & \text{0.09083} & \text{0.02085} & \text{0.00329}  \\
  & \multirow{2}{*}{\it w/ MI} & \textbf{14.95} & \textbf{4.67} & \textbf{1.22} & \textbf{0.0027} & \textbf{0.0117} & \textbf{0.0261} & \textbf{0.0433} & \textbf{0.09100} & \textbf{0.02090} & \textbf{0.00338}\\
  & & (+0.67\%) & (+1.30\%) & (+0.84\%) & (+3.85\%) & (+4.46\%) &(+6.10\%) & (+6.65\%) & (+0.19\%) &(+0.24\%) & (+2.74\%) \\
  \bottomrule
  \end{tabular}
  }
  \caption{Automatic evaluation results for vanilla and MI models on the OpenViDial dataset.}
  \label{tab:automatic}
\end{table}

\subsection{Automatic Evaluation}
We use the following metrics for automatic evaluation:
\begin{itemize}
  \item {\bf BLEU}: Following \cite{li2015diversity,sordoni2015neural}, we report BLEU scores for evaluation. BLEU scores measure the $n$-gram ($n=1,2,4$) overlaps between the generated sequences and gold target sequences. 
  \item  {\bf Diversity}: Following \cite{li2015diversity},
we report the degree of diversity by calculating the number of distinct $n$-grams (Dis-$n$, $n=1,2,3,4$)
in generated responses. The value is scaled by the total
number of generated tokens to avoid favoring long
sentences.  
    \item  {\bf ROUGE-N}: To observe how much information contained in the reference is captured by our model, we report ROUGE-N \cite{lin-2004-rouge}. ROUGE-N is a recall-related measure counting the number of overlapping units based on $n$-grams($n=1,2,4$) between the generated responses and the reference responses. For this evaluation, we only report F-score.
\end{itemize}

\begin{table}
  \centering
  \small
  \scalebox{1}{
  \begin{tabular}{lccc|ccc} \\\toprule
  {\bf Model} & {\bf No $\%$} & {\bf Unsure $\%$}  & {\bf Yes $\%$}& {\bf No $\%$} & {\bf Unsure $\%$}  & {\bf Yes $\%$}\\\midrule
  & \multicolumn{3}{c}{{\it Without MI}} & \multicolumn{3}{c}{{\it With MI}}\\\hline
  NV & 34.6 & 28.6 & 37.8 & 32.0 & 27.4 & 40.6\\
  CV & 25.7 & 25.9 & 48.4 & 24.8 & 25.3 & 49.9\\
  FV & 24.6 & 25.1  & 50.3 & 23.3 & 25.6  & 51.1\\
  \bottomrule
  \end{tabular}}
  \caption{Human evaluation results.}
  \label{tab:human}
\end{table}

\begin{table*}[!t]
  \centering
  \small
  \begin{tabular}{p{4cm}p{4cm}p{5cm}} \\\toprule
  {\bf Time Step $t-2$} & {\bf Time Step $t-1$} &{\bf Prediction (Time Step $t$)} \\\midrule
      \includegraphics[width=90pt,height=60pt]{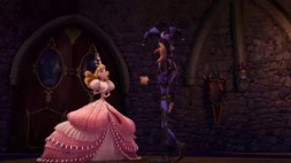} & 
    \includegraphics[width=90pt,height=60pt]{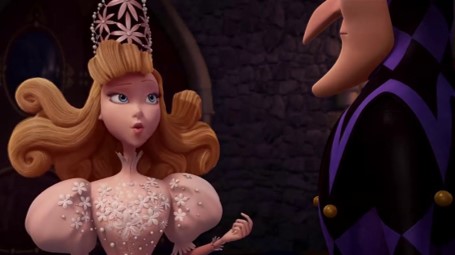} & 
    \includegraphics[width=90pt,height=60pt]{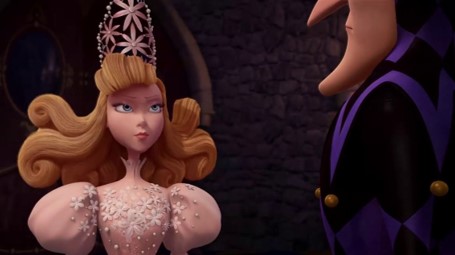}  \\
    {\it Context}: No. &  {\it Context}: The Land of OZ is already falling into ruin because of you. & {\it NV}: I'm sorry. I'm sorry.\\
    &  & {\it CV}: I'm not. I'm not.\\
    &  & {\it FV}: But I'm not a princess.\\
    &  & {\it FV+MI}: I'm not a princess. I can't do it.\\
    &  & {\it Truth}: No good can come from the reign of a fool.\\\midrule
    \includegraphics[width=90pt,height=60pt]{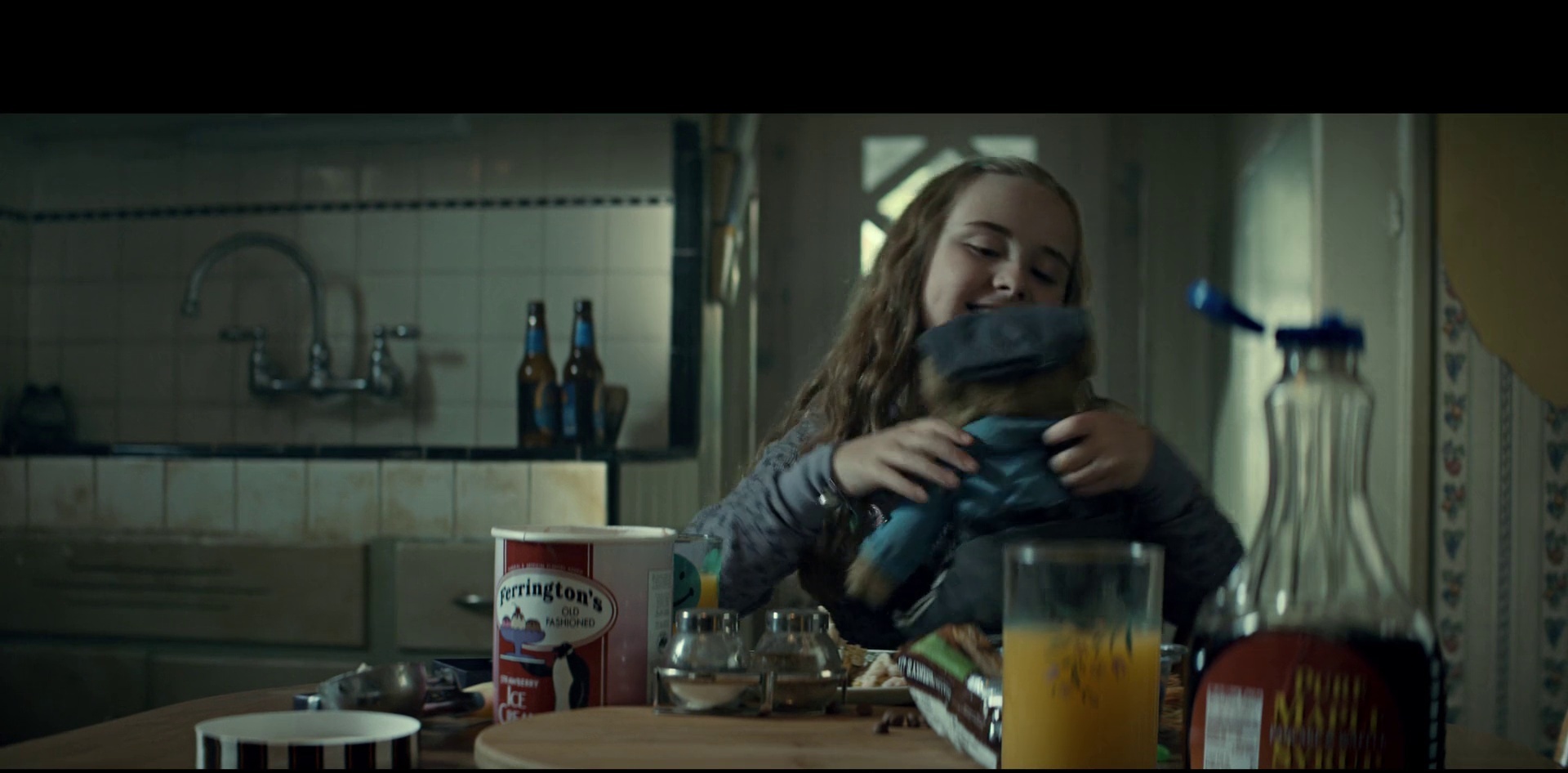} & \includegraphics[width=90pt,height=60pt]{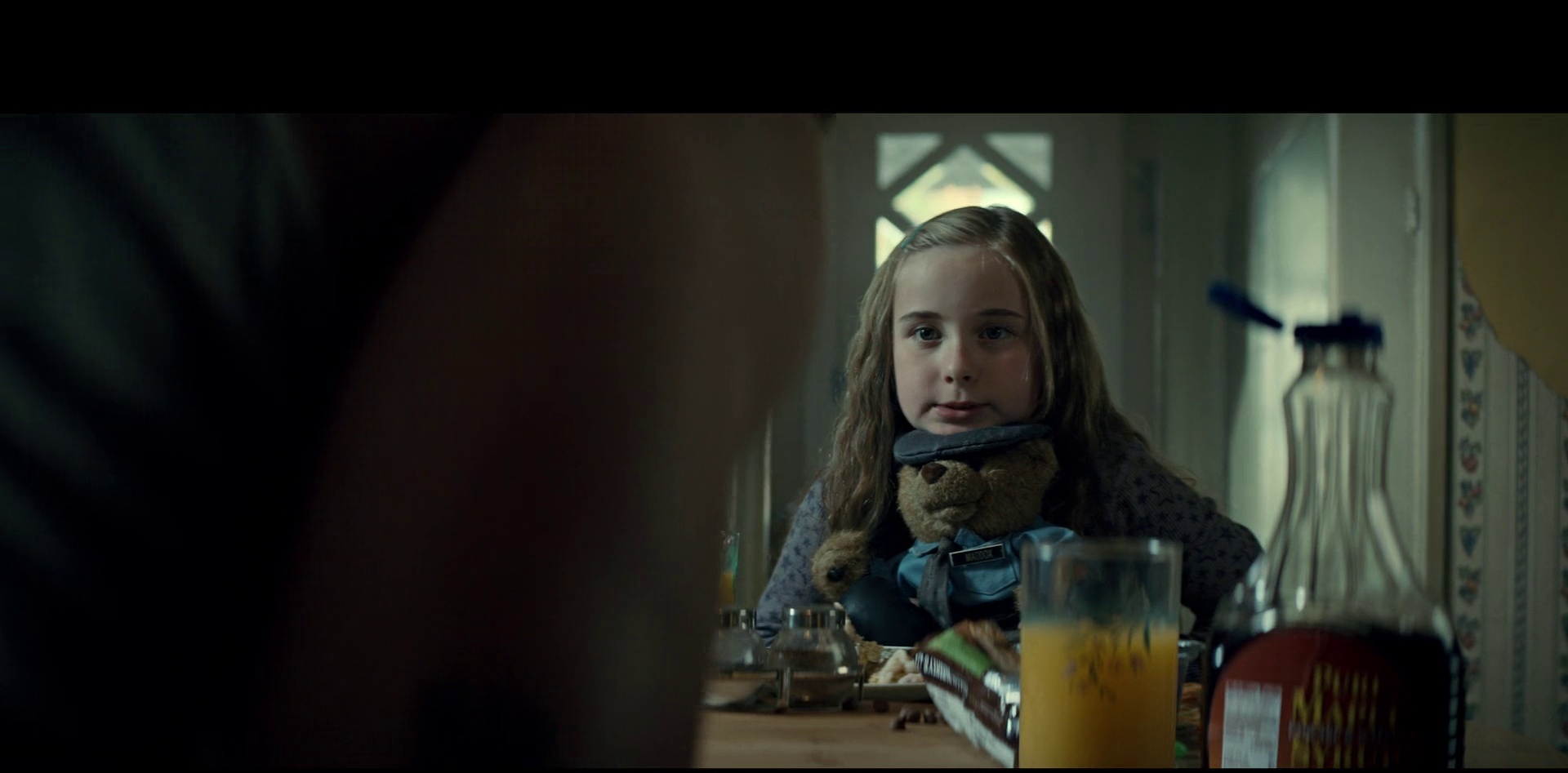} & \includegraphics[width=90pt,height=60pt]{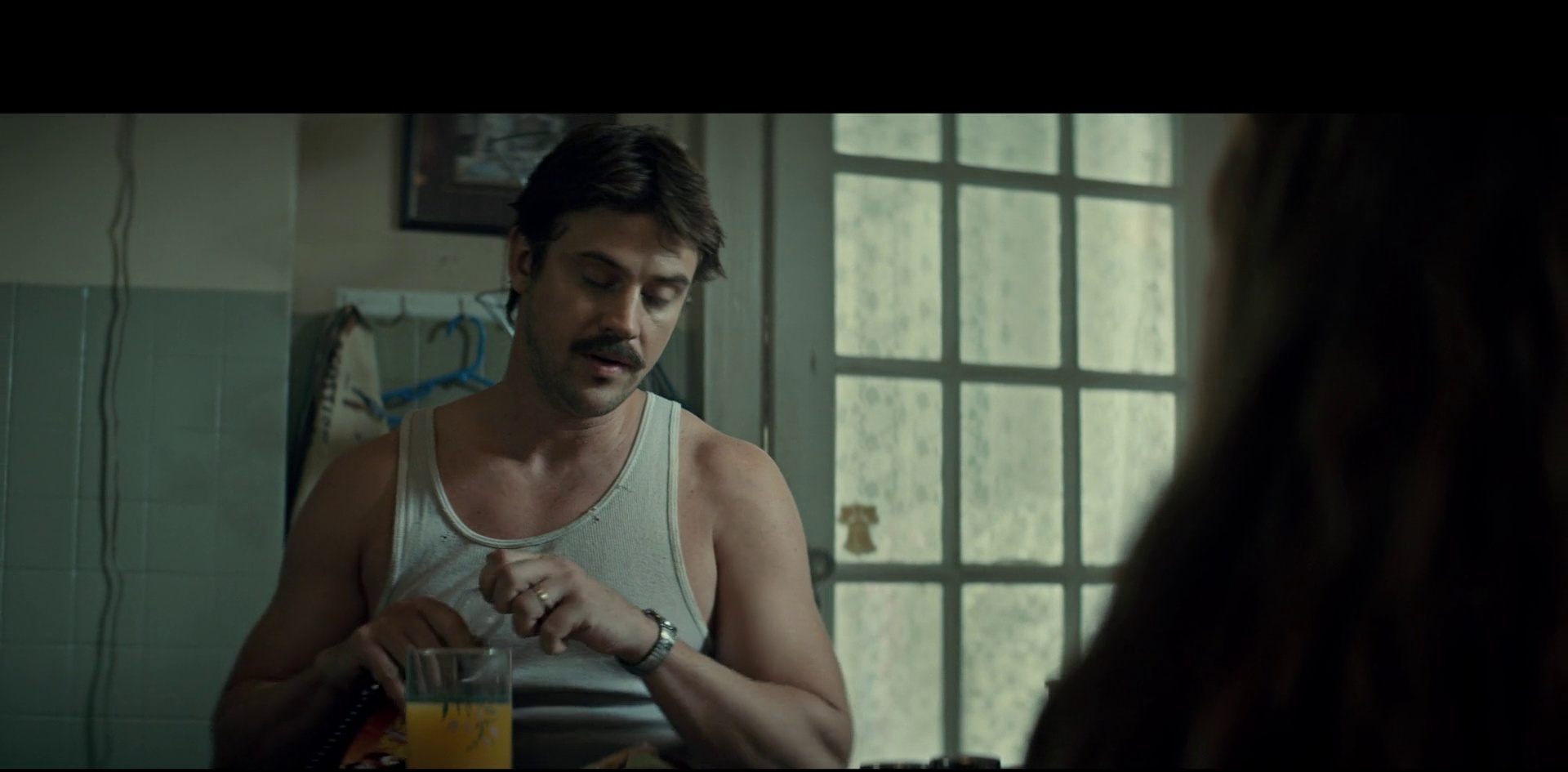}  \\
    {\it Context}: Happy birthday, honey. &  {\it Context}: Bears love bunnies. & {\it NV}: No, no, no, no, no, no.\\
    &  & {\it CV}: Oh, my God.\\
    &  & {\it FV}: I'm gonna take care of it.\\
    &  & {\it FV+MI}: I'm gonna go get some coffee.\\
    &  & {\it Truth}: We need to hurry up.\\\midrule
    \includegraphics[width=90pt,height=60pt]{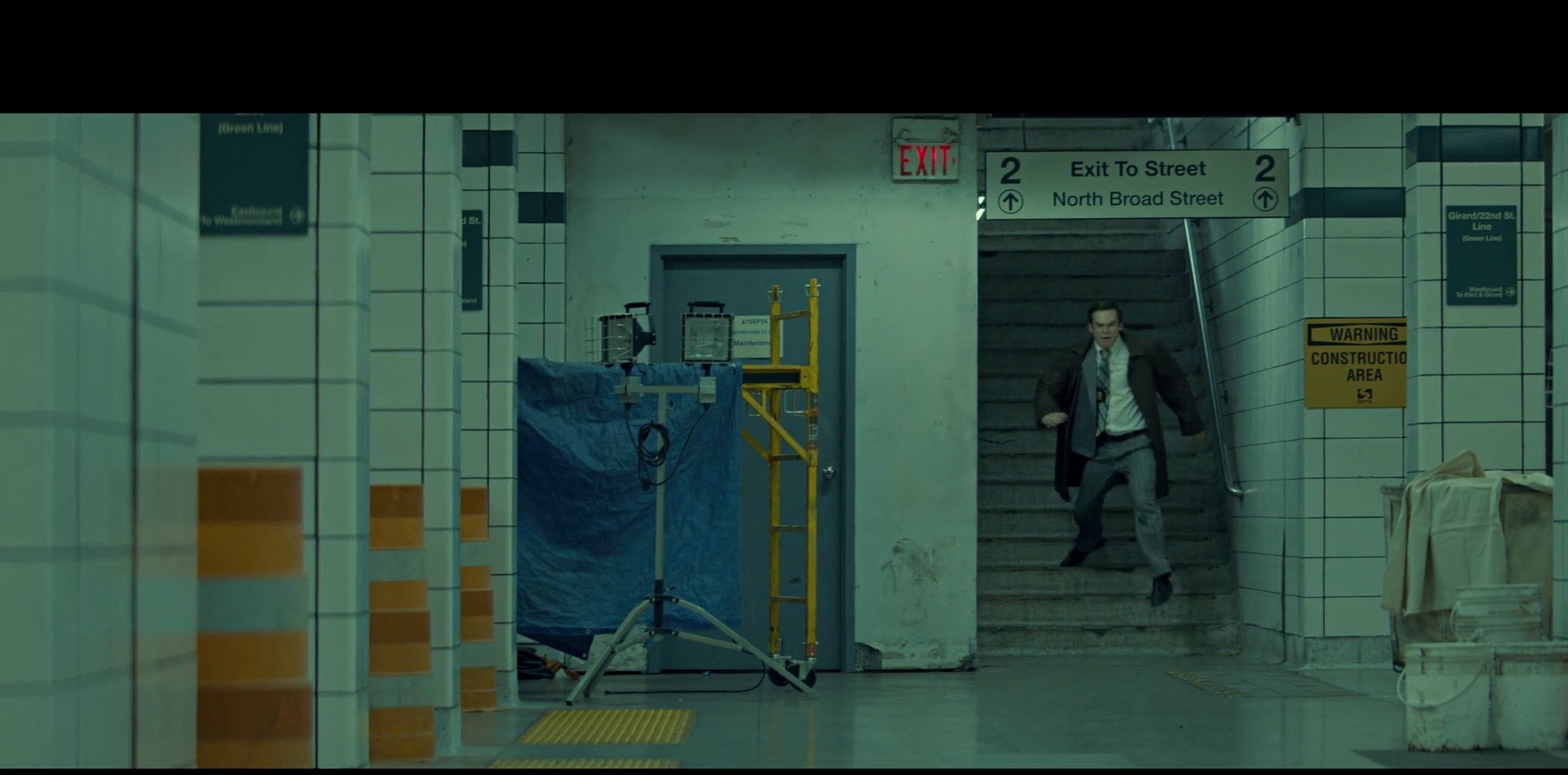} & \includegraphics[width=90pt,height=60pt]{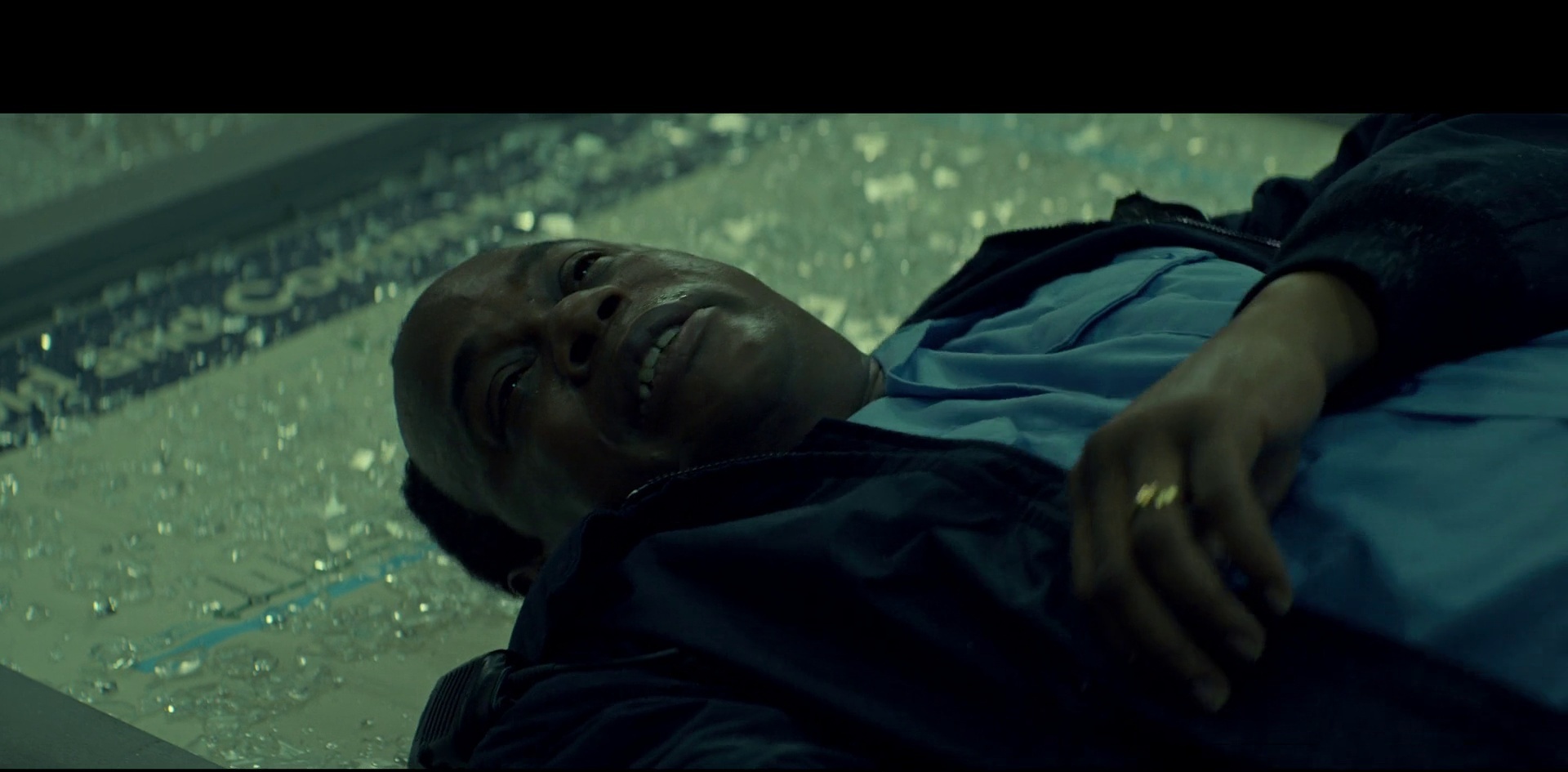} & \includegraphics[width=90pt,height=60pt]{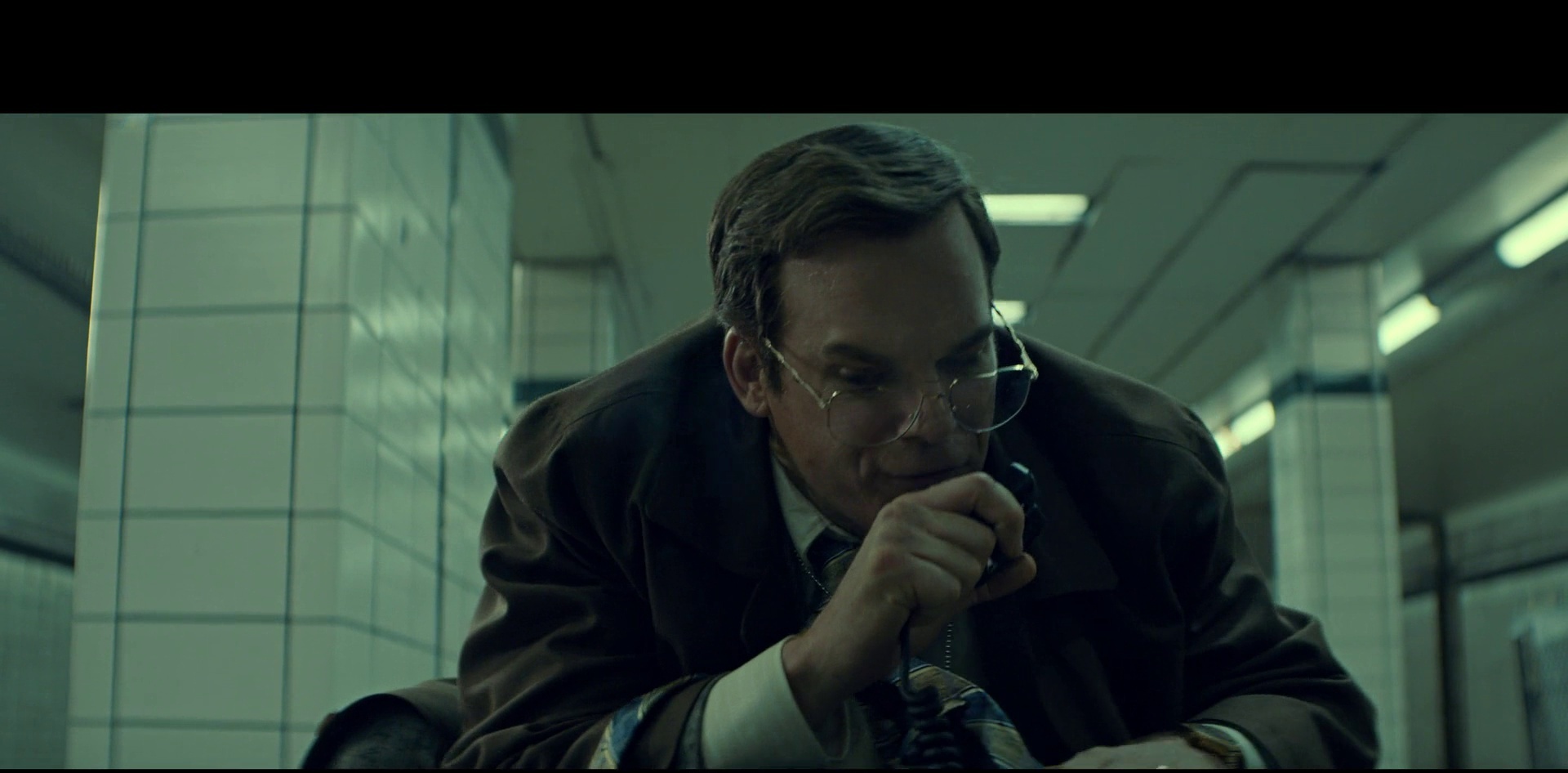}  \\
    {\it Context}: Go! Lock!&  {\it Context}: Officer down. Officer down. & {\it NV}: No, no, no, no.\\
    &  & {\it CV}: Get out of the way!\\
    &  & {\it FV}: Hey, hey, hey!\\
    &  & {\it FV+MI}: I'm on the phone!\\
    &  & {\it Truth}: I need an ambulance at the Girard Street subway.\\
  \bottomrule
  \end{tabular}
  \caption{Examples from the test set generated by vanilla and MI models.}
  \label{tab:visualExample}
\end{table*}

Results are shown in Table \ref{tab:automatic_0} and Table \ref{tab:automatic}. For vanilla visual dialog models, we observe progressive performance boosts from NoVisual to FineVisual along with the increase of fine-grained visual features, indicating that integrating more fine-grained visual features leads to better multi-modal dialog learning abilities. We observe further improvements on those evaluations(i.e., BLEU, diversity, ROUGE) for all three models, i.e., NV, CV, FV, when building mutual dependencies between visual contexts and text utterances.
For example, the BLEU-4 score increases from 1.19 to 1.22 for the FV model when adding the MI component, and the Dis-4 score increases from 0.0406 to 0.0433, with an increment of 6.65\%, the ROUGE-4 also increases from 0.00329 to 0.00338 getting an increment of 2.74\%.
These observations illustrate the effectiveness of modeling fine-grained visual features and visual-text mutual dependency in multi-modal dialog generation.


\subsection{Adversarial Evaluation}
The adversarial evaluation strategy is proposed by \citet{kannan2017adversarial,li2017adversarial}
to 
train a discriminator function to label dialogs as
machine-generated (negative) or human-generated
(positive). Positive examples are taken from
training dialogs, while negative examples are
decoded from a model.
The input to the discriminator is the concatenation of features for  constituent dialog turns, including the preceding features and 
the generated text. 
For each dialog turn, the feature includes both visual features extracted from the image using CNNs, 
and text features using word embeddings. 
A multi-layer 
transformer is built on top of the concatenation, with the \texttt{[CLS]} feature fed to the sigmoid function, the output of which denotes the probability  
of whether the generated text is machine-generated or human-generated. 
We used examples from the dev set to train the discriminator, in which we treat half of the  examples with original responses in the dataset as human-generated, 
and the other half with model generated responses as machine-generated.
We test the trained model on the test set generated in the same way. We report
adversarial success, which is the percentage of the generated responses that can fool the evaluator to believe that it is human-generated.
Higher values of adversarial success indicate better dialog generation models. 
The vanilla NV, CV and FV models respectively obtain adversarial success values of 0.942, 0.917 and 0.890, demonstrating that integrating visual contexts facilitate generating responses more mimicking human conversations.
Further, when combining with MI, MI+NV, MI+CV and MI+FV respectively obtain adversarial success values of 0.915, 0.904, 0.877, showing the advantages  of the MI strategy over its vanilla correspondence.

\subsection{Human Evaluation}
Both automatic evaluations and adversarial evaluation suffer from disadvantages.
For the former, 
there have been debates on their validity  for dialog generation \cite{lowe2018automatic};
for the latter, it requires training another model (i.e., the discriminator) for evaluation. 
We thus conduct human evaluation for further validations. 
We employ crowdsourced judges to provide evaluations for a random
sample of 1000 episodes from the test set. 
For each input context, we present annotators with both preceding text contexts, preceding visual contexts, and the current visual context, 
along with outputs from the three models, i.e., NV, CV and FV.
Annotators were asked to 
 score every model response 
on a 5-point scale (Strongly Agree, Agree, Unsure,
Disagree, Strongly Disagree)
based on three aspects:
{\it Relevance} (whether the
generated response is relevant to the contexts, both visual and textual), {\it Diversity} (whether
the generated response has diverse words) and {\it Readability} (whether the generated response is grammatical). 
Ratings were later
collapsed to 3 categories (Agree, Unsure, Disagree). 
Results are shown in Table \ref{tab:human}.
To verify the statistical significance of the reported results, we perform a pairwise bootstrap test  \cite{johnson2001introduction,berg2012empirical}
to compare the difference between the percentage of responses that are labeled as ``yes''.
We find that FV is significantly better than CV, which is significantly better than NV, with  $p$-value $<$ 0.01.
This validates the importance of harnessing visual contexts for dialog generation.
The MI strategy outperforms its vanilla counterpart also with  $p$-value $<$ 0.01,   
illustrating the effectiveness of modeling visual-text correspondence in dialog generation.

\subsection{Examples}
We randomly choose three examples from the test set and compare the responses generated by vanilla models and the FV+MI model. 
Results are shown in Table \ref{tab:visualExample}.
These examples show that the NoVisual (NV) model and the CoarseVisual (CV) model tend to generate dull and meaningless responses because they discard some salient visual information needed to produce meaningful and context-related sentences. On the contrary, the other two  models -- FineVisual (FV) and FV+MI -- can generate informative responses specific to the visual contexts. 

\section{Conclusion}
In this paper, we propose frameworks to resolve the task of multi-modal dialog generation based on the newly released OpenViDial dataset. Specifically, we propose to model the mutual dependency between text-visual features, leading to the generation of dialog utterances specific to the visual contexts. 
We show that incorporating more fine-grained visual features and integrating MI into multi-modal dialog generation models bring performance improvements regarding both automatic evaluation, adversarial evaluation and human evaluation.

\bibliography{custom}
\bibliographystyle{plainnat}

\end{document}